\pdfoutput=1

\documentclass[11pt]{article}

\usepackage{EMNLP2023}

\usepackage{times}
\usepackage{latexsym}
\usepackage{xcolor}
\usepackage{xspace}
\usepackage{wrapfig}
\usepackage{booktabs}
\usepackage[T1]{fontenc}

\usepackage[utf8]{inputenc}

\usepackage{microtype}

\usepackage{inconsolata}
\usepackage{multirow}
\usepackage{graphicx}
\usepackage{subfig}
\usepackage{amsmath}

\newcommand{\swig}{SWiG\xspace}
\newcommand{\mmee}[0]{M$^2$E$^2$\xspace}
\newcommand{\name}{GenEARL}

\title{\name{}: A Training-Free Generative Framework for Multimodal Event Argument Role Labeling}

\author{
Hritik Bansal, 
Po-Nien Kung, 
\textbf{P. Jeffrey Brantingham}, 
\textbf{Kai-Wei Chang}, 
\textbf{Nanyun Peng}
\\
University of California Los Angeles \\
\texttt{hbansal@cs.ucla.edu, ponienkung@cs.ucla.edu}
}

\begin{document}
\maketitle

\begin{abstract}
Multimodal event argument role labeling (EARL), a task that assigns a role for each event participant (object) in an image is a complex challenge. It requires reasoning over the entire image, the depicted event, and the interactions between various objects participating in the event. 
Existing models heavily rely on high-quality event-annotated training data to understand the event semantics and structures, and they fail to generalize to new event types and domains. 
In this paper, we propose \name{}, a training-free generative framework that harness the power of the modern generative models to understand event task descriptions given image contexts to perform the EARL task. Specifically, \name{} comprises two stages of generative prompting with a frozen vision-language model (VLM) and a frozen large language model (LLM). First, a generative VLM learns the semantics of the event argument roles and generates event-centric object descriptions based on the image. Subsequently, a LLM is prompted with the generated object descriptions with a predefined template for EARL (i.e., assign an object with an event argument role). We show that \name{} outperforms the contrastive pretraining (CLIP) baseline by 9.4\% and 14.2\% accuracy for zero-shot EARL on the \mmee and \swig datasets, respectively. In addition, we outperform CLIP-Event by $22\%$ precision on \mmee dataset. The framework also allows flexible adaptation and generalization to unseen domains.

\end{abstract}

\section{Introduction}

Multimodal event extraction (EE) aims at identifying the event types depicted in an image (e.g., `Arrest'), the event participates (objects), and event argument role labeling (EARL) for the objects (e.g., A person is the `Agent' performing an `Arrest' event). 
In many real-world applications such as analysis of news articles and social media, important event-related information is grounded in the multimodal image-text data \cite{stephens1998rise,blandfort2019multimodal}. For example, a tweet about a protest may not mention whether the protest is violent, which can be readily discerned from the accompanying image \cite{giorgi2022twitter}. Hence, it becomes increasingly important to extract the multimodal events and relations, beyond text, as they provide a useful lens for analyzing social events. 

\begin{figure}[t]
    \centering
    \includegraphics[scale=0.45]{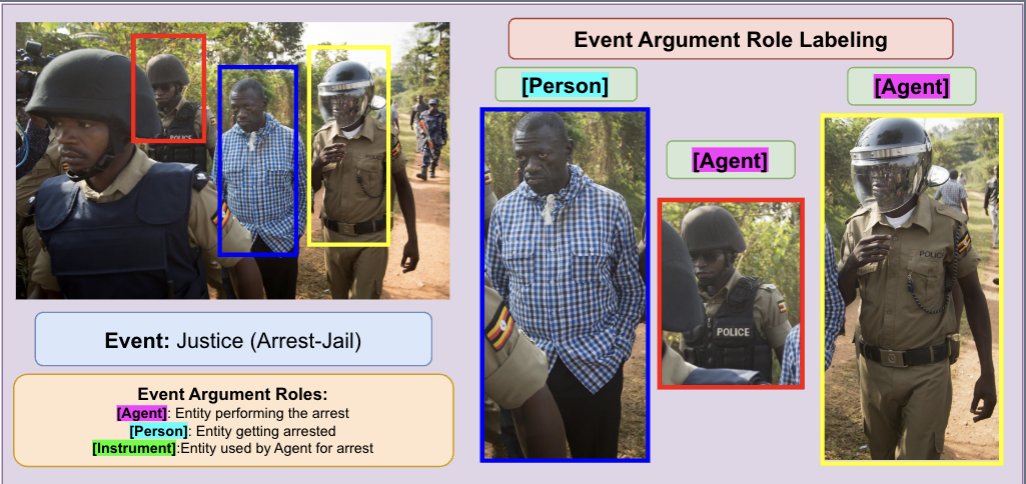}
    \caption{Overview of Multimodal Event Argument Role Labeling task. Given the image that depicts an \textsc{Arrest} event type, a list of possible event argument roles (\textsc{Agent}, \textsc{Person}, \textsc{Instrument}), and three participant objects (bounding boxes) of color `red', `blue', and `yellow'. The task is to assign an event argument to each of the objects based on their role in the depicted event. Here, the `blue' bounding box plays the role of the \textsc{Person} who gets arrested whereas the object in the `red' and `yellow' plays the role of the \textsc{Agent} performing the arrest.\looseness=-1}    
\label{fig:example}
\end{figure}

In this work, our focus is on multimodal Event Argument Role Labeling (EARL)  (illustrated in Figure \ref{fig:example}) given its practical significance and the unique technical challenges it presents within the multimodal EE pipeline. 
For example, social scientists will benefit from understanding the \textsc{Demonstrators}, \textsc{Police}, and \textsc{Instrument} from a \textsc{Protest} scene. 
In the meanwhile, performing an accurate multimodal EARL requires fine-grained understanding of the entire image depicting the event along with comprehending the roles and interactions between various objects present in the image, 
which poses challenges for large-scale pretrained vision-language models such as CLIP \cite{radford2019language}. 

Existing methods \cite{li2020cross,li2022clip} 
trained on event-annotated data, such as ACE \cite{doddington2004automatic}, imSitu \cite{yatskar2016situation}, and VOA  \cite{li2020cross} tend to overfit to the event types seen during training, and fail to generalize well to unseen event types and new domains \cite{parekh2023geneva}. 
Subsequent finetuning of these models will require the creation of new event-annotated data, which is expensive and hard to acquire. 

To tackle these challenges, our major contribution is the proposal of \textbf{\name{}}, a two-stage training-free generative framework for multimodal EARL. Our framework utilizes the ability of the instruction-following generative vision-language models (GVLM) and large language models (LLM) to effectively comprehend multimodal and text-only input prompts. First, we prompt the GVLM with visual features of the image and object, along with the target event and its associated argument role in a predefined template such that it generates an event-centric object description. This object description summarizes the role of the object in the context of the event depicted in the image. Since the generative VLM reasons over the task description in the input prompt, it does not require expensive event-annotated data for generating object role descriptions. In the next stage, we prompt instruction-following LLM with the generated object descriptions, image caption, along with the targeted event and its argument role labels in a predefined template to predict event argument role labels. 
We further show that 
in-context learning \cite{dong2022survey,liu2023summary} for each of the steps can improve EARL extraction. 

\begin{figure*}[h]
    \centering
    \includegraphics[scale = 0.4]{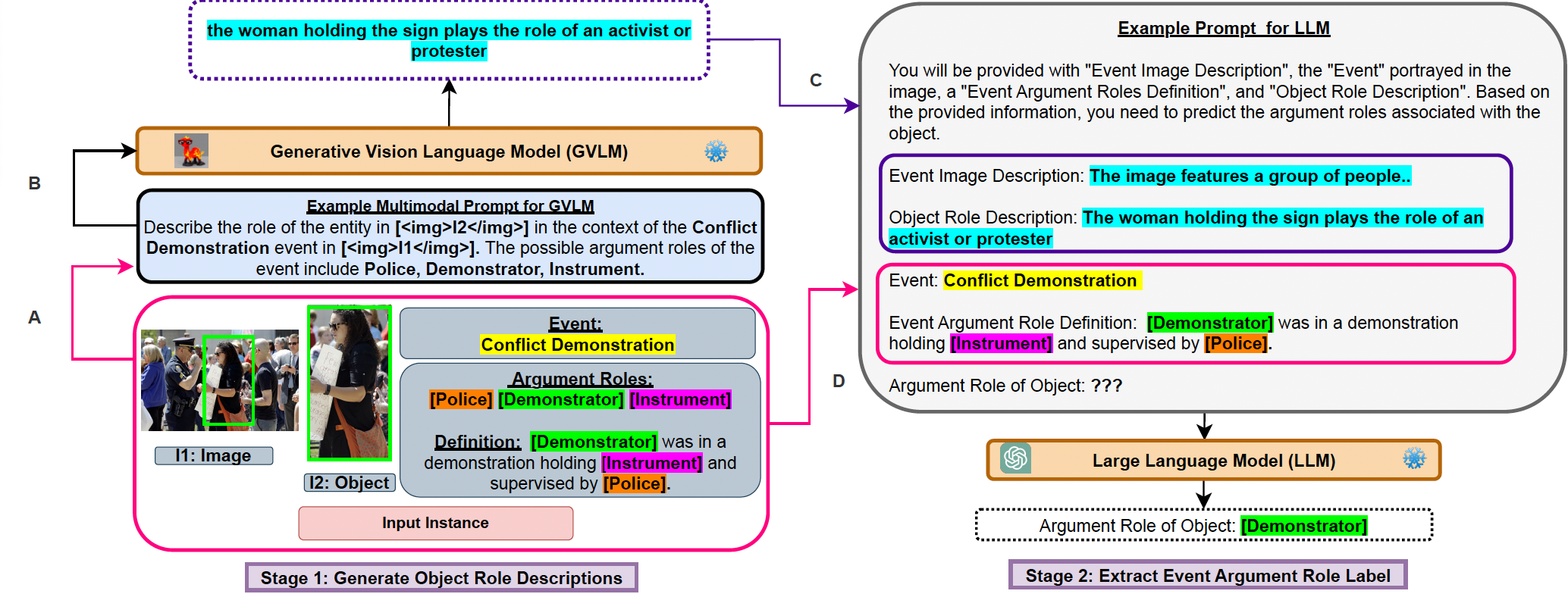}
    \caption{Overview of the \name{} framework for multimodal event argument role labeling. It comprises two stages of generative prompting. In the first stage, a generative vision-language model like LLaVA is prompted with the multimodal input prompt to generate an event-centric object role description. In the second stage, we extract the argument role label based on the generated object description using a large language model like ChatGPT. \textbf{A:} The raw input instance data is converted into a multimodal prompt for GVLM. \textbf{B:} The input prompt is fed to the GVLM for object role description generation. \textbf{C:} The generated object role description is embedded in the template used to prompt the LLM. GVLM is also used to generate a caption for the image that depicts the event. \textbf{D:} The event details including the possible argument role labels are added in the LLM prompt.\looseness=-1}    \label{fig:genearl}
\end{figure*}

Our second contribution is a comprehensive empirical study of the generalization ability of our framework to perform multimodal EARL on the \mmee and \swig datasets, as used in \citet{li2022clip}. We find that \name{} outperforms the CLIP-L/14 \cite{radford2019language} baseline by $9.4\%$ and $14.2\%$ accuracy in the zero-shot EARL on the \mmee \cite{li2020cross} and \swig \cite{pratt2020grounded} datasets, respectively (\S \ref{exp:earl}). In addition, we find that combining CLIP for event detection and \name{} for EARL outperforms CLIP-Event \cite{lu2021text2event} on \mmee by 22\% on precision.

We further study the discrepancies between the training-free and supervised paradigms by finetuning a CLIP-B/32 on the training data of the \swig dataset (\S \ref{setup:baseline}), which enhances the CLIP's understanding of event semantics and structures. We observe that our framework reduces the gap between the supervised and training-free paradigms from $44.5\%$ to $16.1\%$ on the \swig dataset. 
In addition, our framework outperforms this finetuned CLIP model on the \mmee dataset by $3\%$ accuracy with just three-shot in-context learning examples. This highlights that our generative paradigm is competitive with supervised methods 
without any training on event data. 

\section{Background: Multimodal EARL}
\label{sec:background}
In this section, we first describe the task of multimodal event argument role labeling (EARL). 
Given an input image $\mathcal{I}$ that is depicting an event type $E$. The event type belongs to one of the possible events types $E \in \mathcal{E}$ under the predefined event ontology. Further, each event type $E$ consists of a set of event arguments $\mathcal{A}_E = \{a_{1},\ldots,a_{m}\}$ that specifies the roles associated with it. In our datasets, each image contains a set of objects $\mathcal{O} = \{o_1,\ldots,o_n\}$ where each object $o_i$ is represented as a bounding box and participates in the depicted event. 

The task of multimodal event argument role labeling (EARL) is to label every participating object (bounding box) $o_{i}$ with an event argument role label $a_{j}$. The predicted argument role label for the participating object highlights its interactions with the other objects in the scene and its contribution to the understanding of the event scenarios. For example, Figure \ref{fig:example} the object in the `blue' bounding box plays the role of the \textsc{Person} who gets arrested whereas the object in the `red' and `yellow' plays the role of the \textsc{Agent} performing the arrest. Overall, EARL is a challenging task since it requires a model to reason over the fine-grained features of the image and the interactions between the participating objects. We highlight the assumptions of our work in Appendix \S \ref{app:method_earl}. 

\section{The \name{} Model}
\label{Method}

To tackle the challenging task of multimodal EARL, we propose \name{}, a two-stage training-free generative framework that utilizes the unique capabilities of the modern generative models \cite{liu2023visual,chatgpt} to understand event task descriptions and perform well on them. Specifically, the framework consists frozen (a) generative vision-language model (GVLM) and (b) large language model (LLM). Consider an input instance $\mathcal{Q}_i = \{\mathcal{I}, E, o_i, \mathcal{A}_E\}$ where we aim to label the event argument role for the object $o_i$ in the event $E$ depicted in the image $\mathcal{I}$. $\mathcal{A}_E$ comprises the possible event argument role labels for the event $E$ along with their definition from the dataset guidelines. 

We propose to decouple the task of multimodal EARL on the given input instance into two distinct sub-tasks. This approach allows each sub-task to be effectively addressed by the individual capabilities of the GVLM and LLM respectively. Specifically, we (a) generate an event-centric object description that captures the role played by the object $o_i$ in the context of the event $E$ in the image $\mathcal{I}$ (\S \ref{method:gvlm}) followed by (b) extracting the EARL from the generated object role descriptions (\S \ref{method:llm}).

\subsection{Generative Vision-Language Model}
\label{method:gvlm} 

We consider a generative vision-language model (GVLM) $\mathcal{G}$ that can reason over the text and visual input and generate an output text. Here, the GVLM is capable of capturing the fine-grained details in the event task description with image(s) in its context. We illustrate this stage in Figure \ref{fig:genearl} (left).

For the input instance $\mathcal{Q}_i$ we prompt the GVLM with a multimodal prompt $\mathcal{P} = \mathcal{T}(\mathcal{Q}_i)$ where $\mathcal{T}$ is a predefined template that converts the input query details into a multimodal prompt (Appendix Table \ref{tab:llava_main}). We highlight the example multimodal input prompt in Figure \ref{fig:genearl} (left). Subsequently, we generate the event-centric object role description $d_{o_i}$ from the GLVM with the context $\mathcal{P}$.

\subsection{Large Language Model}
\label{method:llm}

We illustrate this stage in Figure \ref{fig:genearl} (right). In the first stage, we generate an object role description for a given image $\mathcal{I}$ and participant object $o_i$. Here, we us a large language model (LLM) $\mathcal{L}$ since they are capable of generalizing to novel input queries to solve a task \cite{wei2021finetuned}. We provide the LLM with a text-only prompt $\mathcal{C} = \mathcal{T}_{\ell}({d_{\mathcal{I}}, E, \mathcal{A}_E, d_{o_i}})$. Here, $\mathcal{T}_{\ell}$ is a predefined template that converts the input information including the generated object role description $d_{o_i}$ from the previous stage, event $E$, image description $d_{\mathcal{I}}$ \footnote{ $d_{\mathcal{I}}$ is the image caption generated using the GVLM $\mathcal{G}$ with the prompt `Describe this image in detail.'.}, and the possible argument roles $\mathcal{A}_E$ to detailed task description for the LLM. We illustrate an example input prompt at the top of Figure \ref{fig:genearl} (right). 

The template $\mathcal{T}_\ell$ is designed in a way that prompts the LLM to output the predicted event argument role $\tilde{a}_i$ for the object $o_i$. Following this, we can evaluate if the predicted EARL matches with the ground-truth argument role label $a_i$. We show the example predicted EARL in the lower half of Figure \ref{fig:genearl} (right). In our work, we use ChatGPT \cite{chatgpt}, a state-of-the-art LLM that can adapt to new tasks and perform well on them. 


\section{Experimental Setup}
\label{setup}

\subsection{Dataset}

We evaluate our approach on two datasets: \mmee \cite{li2020cross} and \swig \cite{yatskar2016situation}. In both these datasets, each image has an associated event type, a bounding box for the participating objects, and their associated event argument roles.

\mmee consists of 391 images with event mentions taken from the Voice of America (VOA) news website\footnote{\url{https://www.voanews.com/}} where the event types and their argument roles are derived from the ACE ontology \cite{doddington2004automatic}. To maintain a good argument role density for evaluation, we filter the event types with less than three argument roles (Appendix Table \ref{m2e2_event_list}). In total, we label 990 bounding boxes spanning 275 images in the \mmee dataset. 

\swig dataset consists of images that can belong to one of 504 action verbs. Following \citet{li2022clip}, we consider the action verbs to be analogous to the event types. Like our \mmee setup, we filter the action verbs with less than three event argument roles (Appendix Table \ref{swig_event_list}). In our experiments, we label 1600 bounding boxes spanning 600 test images from this dataset. 

Following \citet{li2022clip}, we do not consider `place' as an event argument role since it cannot be grounded as a bounding box for both datasets. In the zero-shot setting, we evaluate the \name{}'s capability to reason about the examples, without training on event data, by prompting the pretrained generative models with event information, its associated argument role labels, and the visual context from the image and the object. In the few-shot setting, we provide the k-shot examples for all the bounding boxes in k images sampled randomly from the dataset. For example, in case k = 1, we show the labels for the objects in one of the images from the dataset (Appendix Table \ref{tab:chatgpt_one_shot}).

 \begin{table*}[h]
\begin{center}
\resizebox{0.7\linewidth}{!}{%
\begin{tabular}{llcc}
\hline
\textbf{Paradigm}    &   \textbf{Method}   & \textbf{\mmee Accuracy (\%)} & \textbf{\swig  Accuracy (\%)}  \\\hline
\multirow{5}{*}{Training Free} &CLIP-B/32 (0-shot)   &   28.1                 &     22.6                     \\
 & CLIP-L/14 (0-shot) &  30                  &         24.7                    \\
 & \name{} (0-shot)    & 39.4               & 38.9                            \\
& \name{} (1-shot)      & 44.1               & 50.1                          \\
& \name{} (3-shot)      & \textbf{45.8}                & \textbf{51}  \\
\hline
\textcolor{gray}{Supervised}             & \textcolor{gray}{CLIP-B/32 (+ \swig)}       &  \textcolor{gray}{41.9} &  \textcolor{gray}{67.1}  \\\hline 
\end{tabular}%
}
\end{center}
\caption{Multimodal EARL accuracy (\%) of \name{} (zero-shot to three-shot), pretrained CLIP (zero-shot), and supervised CLIP on the \mmee and \swig dataset. Zero-shot \name{} framework outperforms the zero-shot CLIP models on both datasets. The performance of our framework improves from zero-shot to three-shot example without any training of the underlying models on the event-annotated data. A supervised baseline is only used as a reference as it accesses event-annotated data. We present precision, recall, and F1 scores for the same in Appendix \S \ref{sec:p_r_f1}. \looseness=-1}
\label{tab:main_earl_results}
\end{table*}

\subsection{Implementation Details}

\name{} is a two-stage framework that consists of a vision-language model and a large language model. In our experiments, we use LLaVA-7B \cite{liu2023visual}, a state-of-the-art generative vision-language model. It is finetuned with 150K multimodal instructions that enable it to understand a wide breadth of concepts and generate helpful and coherent outputs. Traditionally, LLaVa is used for generating outputs for a single image and text inputs, however, we adapt it reason over multiple images during inference i.e., an image depicting the event and a bounding box containing the object to be labeled. Due to this design, the multimodal input prompt can flexibly incorporate the input instance details to generate the event-centric object descriptions with any training on the event-annotated data. Existing models \cite{li2022clip} cannot adapt to new event types or new domains without further finetuning on an event-annotated dataset.

For the second-stage of our framework where we predict the event argument role label based on the generated event-centric object description, we use ChatGPT \cite{chatgpt} as our default LLM. It is a powerful LLM that can adapt to novel task descriptions and perform well on them. In addition, it is capable of learning about the target domains through few-shot examples in its context. In our work, we leverage this capability by prompting the LLM with a few-shot solved examples. Since prompting ChatGPT comes with cost considerations \footnote{\url{https://openai.com/pricing}}, we label the event argument roles in batches where all the participating objects $o_i \in \mathcal{O}$ in an event depicting image $\mathcal{I}$ form a single batch. We access this model through OpenAI's API \footnote{\url{https://openai.com/blog/openai-api}} where it corresponds to `gpt-3.5-turbo'. In \S \ref{exp:llm}, we use Alpaca \cite{alpaca} by replacing ChatGPT in the second-stage for multimodal EARL.

\subsection{Baselines}
\label{setup:baseline}

We compare the \name{} framework against the open-source CLIP models \cite{radford2019language}. We evaluate our framework against zero-shot CLIP to understand the generalization capability of our model without training on any event-annotated data. Subsequently, we train a supervised CLIP model to assess the assess the gap between our training-free paradigm and a supervised model that learns about the events from event annotated data.

Following \citet{li2020cross}, we compare against the large-scale representation learning vision-language models like \textbf{CLIP-B/32} and \textbf{CLIP-L/14} \cite{radford2019language} since they can leverage any event knowledge acquired during its pretraining for multimodal EARL in a zero-shot setting. 

Consider an event $E$ depicted in an image, a set of argument role labels $\mathcal{A}_E = \{a_1,\ldots, a_m\}$ for the event, and the object $o_i$ that needs to be labeled. We calculate the similarity score $s(o_i, T_K(a_j))$ between the object representation in the image embedding space and the text embedding of a set of text-based templates $T_K$ containing information about the event and event argument role. Specifically, our template will be `An object playing $a_j$ role in the $E$ event.' for every argument role label $a_j$ in $\mathcal{A}_E$'. The predicted argument role label $\tilde{a}_i$ will be the one that maximizes the similarity score between the object representation and its corresponding text template i.e., $\tilde{a}_i = \text{argmax}_{j}(s(o_i, T_K(a_j))$. 

To impart the knowledge of the event semantics and structures to the pretrained baseline, we find a CLIP-B/32 model on the 75K images of the \swig dataset. We create a text-based prompt with the event and argument role label for every example, as describe above. Finally, the model is finetuned with the contrastive pretraining objective to bring the representations of the objects and their EARL together in the joint embedding space and pull apart the representations of the unmatched objects and event roles. We provide the finetuning details in Appendix \S \ref{app:clip_supervised}. Given the lack of training data for the \mmee domain, we conduct an evaluation of \textbf{CLIP (+\swig)} as a transfer learning setup.

\section{Experimental Results}

\begin{table*}[h]
\begin{center}
\resizebox{0.8\linewidth}{!}{%
\begin{tabular}{lccc}
\hline
      \textbf{Method}  &   \textbf{\mmee Accuracy (\%)} & \textbf{\swig Accuracy  (\%)} & \textbf{Average (\%)} \\\hline
\name{} w/ ChatGPT (0-shot)  & 39.4               & 38.9               & 39.2                  \\
\name{}  w/ Alpaca-7B (0-shot) & 50.4              & 36.5               & 43.5                \\
\name{} w/ ChatGPT (1-shot) & 44.1               &\textbf{50.1}               & \textbf{47.1}                  \\
\name{} w/ Alpaca-7B (1-shot) & \textbf{54.6} & 37.5               & 46.1 \\ \hline

LLaVA-7B (0-shot)    & 28.6 &    26                &    27.3                     \\\hline
\end{tabular}%
}
\end{center}
\caption{Multimodal EARL accuracy (\%) of the \name{} framework with the choice of LLM (\S \ref{exp:llm}). The last row highlights the performance if we were to only use the GVLM for multimodal EARL (\S \ref{exp:earl_gvlm}).\looseness=-1}
\label{tab:llm_results}
\end{table*}

\subsection{Main Results}
\label{exp:earl}

We evaluate the accuracy of the \name{} framework for multimodal EARL on the \mmee and \swig datasets, in Table \ref{tab:main_earl_results}. We find that zero-shot \name{} achieves a multimodal EARL accuracy of $39.4\%$ and $38.9\%$ on the \mmee and \swig datasets, respectively, without any specialized training on the event-annotated data. We observe that zero-shot \name{} framework outperforms the CLIP-L/14 baseline by $9.4\%$ on \mmee and $14.2\%$ on \swig dataset. This highlights the strong generalization ability of the generative models to adapt to multimodal EARL task descriptions and perform well on them.

Prior works have shown that the LLM are capable of improving their reasoning capabilities by learning from the few-shot examples in their context \cite{dong2022survey,brown2020language}. Hence, we prompt the ChatGPT model in the \name{} framework with one-shot and three-shot settings. We observe that even a single-shot example in the context greatly enhances the performance of the framework to 44.1\% and 50.1\% on the \mmee and \swig datasets, respectively. In addition, we find that providing three-shot also slightly improves the EARL accuracy over the one-shot setting. Due to the context length limitations and cost considerations, while using the OpenAI API, we do not experiment with more than three-shot examples. Our results highlight the unique capability of our framework to improve its performance with a very small amount of event-annotated data.

We further aim to understand the performance improvements and generalization of the pretrained CLIP baseline when it is finetuned with the \swig training data, which was originally acquired by expensive annotation from the human labelers. We observe that our framework reduces the gap between the supervised paradigm and training-free paradigm from $44.5\%$ to $16.1\%$ on the \swig dataset. In addition,  our framework outperforms the finetuned CLIP model on the \mmee dataset by $3\%$ accuracy with just three-shot examples. Thus, our framework provides a flexible, training-free, and generalizable method for accurate multimodal EARL that can easily adapt to new event types. We provide a few qualitative examples for model predictions in Appendix Figure \ref{fig:qual_example}.

\paragraph{Comparison with CLIP-Event.}

We propose \name{} as a multimodal EARL framework where we assume access to the ground-truth event depicted in the image. This differs from prior work CLIP-Event \cite{li2022clip} that performs event detection followed by EARL. Hence, we use a combination of the CLIP-B/32 for event type detection and \name{} framework for EARL to directly compare with the reported CLIP-Event results.\footnote{\textcolor{red}{The code for CLIP-Event is publicly available, however, its pretraining dataset or pretrained checkpoint are unavailable.}}

\begin{table}[h]
\resizebox{\linewidth}{!}{%
\begin{tabular}{lccc}
\hline
\multirow{2}{*}{\textbf{Method}}      & \multicolumn{3}{c}{\mmee}    \\
                             & \textbf{Precision} & \textbf{Recall} & \textbf{F1}     \\\hline
CLIP-Event     & 21.1\%    & 13.1\%   & 17\%  \\  
CLIP + \name{} (0-shot) & 42.4\%    & 24.1\% & 31.4\% \\
CLIP + \name{} (3-shot) & \textbf{43\%}      & \textbf{29\%}   & \textbf{34.6\%} \\\hline
\end{tabular}%
}
\caption{Comparison between the combination of CLIP (for event detection) and \name{} (for EARL) with CLIP-Event that performs end-to-end event detection followed by EARL.}
\label{sec:comp_clip_event}
\end{table}

In Table \ref{sec:comp_clip_event}, we find that the combination of CLIP and \name{} framework outperforms CLIP-Event on precision and recall by 22\%, 15\%, 17\% on precision, recall, and F1 respectively. With advancements in the event detection modules, the performance of the complete event detection and \name{} pipeline will further improve.


\subsection{The Impact of Large Language Models}
\label{exp:llm}


Here, we aim to study whether ChatGPT can be replaced by a relatively smaller instruction-following LLM like Alpaca-7B \cite{alpaca} in our \name{} framework. We provide the details for predicting EARL using Alpaca in Appendix \S \ref{app:alpaca_earl}.

We compare the EARL accuracy with ChatGPT and Alpaca in the \name{} framework under the zero-shot and one-shot settings in Table \ref{tab:llm_results}. We observe that Alpaca-7B outperforms ChatGPT in the zero-shot setting ($43.5\%$ vs $39.2\%$) while lagging behind ChatGPT in the one-shot setting ($46.1\%$ vs $47.1\%$) in the average EARL accuracy across the datasets.\footnote{We do not perform a three-shot evaluation with Alpaca due to the limitations of its context length and ability to capture long-range dependencies.} It suggests that ChatGPT is able to improve more with the few-shot examples, while the base performance of Alpaca is already high. Specifically, we find that Alpaca-7B outperforms ChatGPT by a large margin on the \mmee dataset suggesting that it understands the events grounded in the news data better than ChatGPT. We observe that ChatGPT outperforms Alpaca-7B by a large margin on the \swig dataset which suggests that ChatGPT is better at reasoning over the event types associated with the common action verbs. Such differences in the understanding of the event domains can be attributed to the difference in the base language models. InstructGPT \cite{ouyang2022training} for ChatGPT and LLaMA \cite{touvron2023llama} for Alpaca, and their pretraining data.

\begin{figure}[h] \subfloat[\centering\label{exp_fig:dist_responses} Distribution of labelers responses \looseness=-1]{{\includegraphics[scale = 0.5]{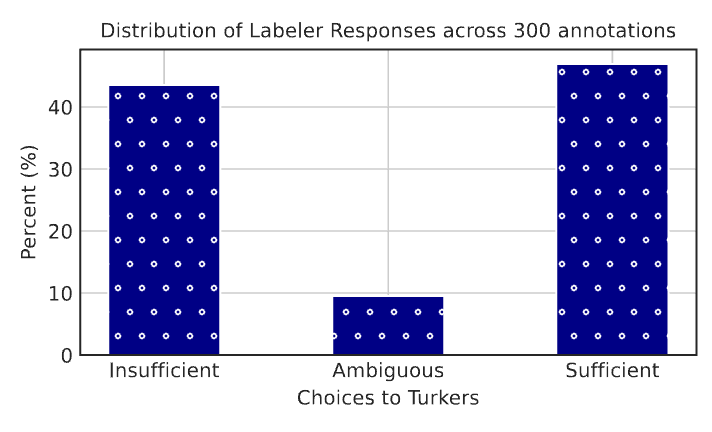}}} \\\subfloat[\centering\label{exp_fig:agreement_responses} Agreement between \name{} prediction and the quality of the object role descriptions assessed by the human labelers.]{{\includegraphics[scale = 0.5]{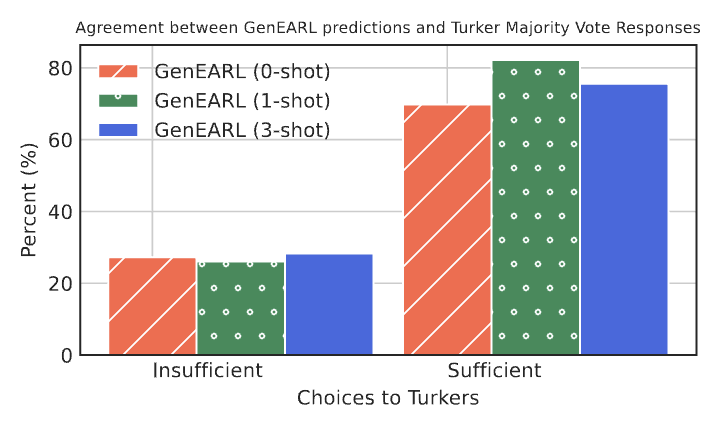}}}
    \caption{Human assessment of the quality of the generated object role descriptions from the LLaVA model.}
    \label{exp_fig:human_assessment}
\end{figure}

\subsection{Human Assessment}
\label{exp:human}

We hypothesize that the \name{}'s performance depends on the quality of the object role descriptions generated by the GVLM (\S \ref{method:gvlm}). To verify this, we perform a human assessment for 100 instances of the generated object descriptions from the \mmee dataset. We ask the human labelers to choose whether the generated object role descriptions provide sufficient information for a human or LLM to correctly label the object's event argument role. We conduct the experiment using 4 labelers from Amazon's Mechanical Turk (MTurk) platform.\footnote{\url{https://www.mturk.com/}} Prior to the annotation, each labeler passed a qualification test, involving 10 sample instances, different from the actual 100 instances, to assess their understanding of the task. We provide the annotators with a detailed slide deck describing the task with a few solved examples. We attach a screenshot for one of the solved examples in Figure \ref{fig:turker_solved}. We assign 3 labelers per instance to evaluate the annotator agreement and get majority votes on the description quality. The annotators were paid at a rate of 18\$ per hour \footnote{The total cost of human annotations was close to \$100.}.

We find that the agreement between any two annotators averaged over 100 instances was $75\%$ which is high given the subjective complexity involved in assessing the dense object role descriptions. In Figure \ref{exp_fig:dist_responses}, we provide the distribution of all the labeler responses for 300 annotations (100 instances $\times$ 3 labelers). We observe that the labelers find $47\%$, $10\%$, $43\%$ of the object role descriptions sufficient, ambiguous, and insufficient, respectively, to correctly label the object's argument role. This suggests that there is a scope for enhancements in the quality of the generated object descriptions with advancements in the GVLM design and prompting techniques. In Figure \ref{exp_fig:agreement_responses}, we assess the performance of the \name{} framework on the instances that were marked containing `insufficient' and `sufficient' information to accurately predict the object role labels. We find that the results verify our hypothesis since the performance of the framework is $\sim 80\%$ and $\sim 20\%$ on the high and low-quality examples, respectively. The result highlights that innovations in the generative VLM will improve the performance of the framework.

\subsection{Ablations}
\label{exp:ablation}

\paragraph{EARL via GVLM}
\label{exp:earl_gvlm}

In our framework, we utilize LLaVA-7B for generating the event-centric object role descriptions that subsequently prompt the LLM for multimodal EARL prediction. Here, we study whether we can perform multimodal EARL without \name{}'s two-stage pipeline. To answer this, we use LLaVA-7B to label the event argument roles for the participating objects given the visual features of the image and object, and the event and its associated argument roles. We provide more details for predicting EARL using LLaVA-7B in Appendix \S \ref{app:llava_earl}.

We present our findings regarding multimodal EARL with LLaVA-7B under the zero-shot setting, as shown in the last row of Table \ref{tab:llm_results}. Our analysis reveals that LLaVA-7B exhibits subpar performance when compared to the \name{} with LLMs on both datasets. The underperformance of LLaVA-7B in the novel event argument role labeling task can potentially be attributed to its fine-tuning data, which lacked classification-oriented instructions and instead emphasized providing detailed image descriptions following the input instruction. Consequently, we leverage LLaVA's capability to generate event-centric object descriptions when prompted, which can be effectively utilized by a robust LLM within the \name{} framework. This LLM is likely to have encountered a greater number of classification-style instructions during its finetuning process which leads to its labeling capabilities.

\paragraph{Effect of the Conditioning Variables}
\label{exp:conditioning}

In the previous experiments, we show that the quality of the generated object role descriptions affects the predictions made by the LLM in the \name{} framework. In this experiment, we aim to study the factors that affect this quality. Specifically, we compare the performance of our framework where object descriptions are generated (a) without the visual features of the event depicting image $\mathcal{I}$, and the event details including the event $E$ and argument role label and their definitions $\mathcal{A}_E$, and (b) without the visual features of the event depicting image $\mathcal{I}$. In both scenarios, the visual features of the object $o_i$ being labeled are present in the input prompt to the GVLM. The input template for these settings are present in Table \ref{tab:llava_wo_image} and Table \ref{tab:llava_wo_image_event}.

\begin{table}[h]
\resizebox{\linewidth}{!}{%
\begin{tabular}{lcc}
\hline
\textbf{GVLM prompting}      & \textbf{\mmee  (\%)} & \textbf{\swig  (\%)}  \\\hline
 w/ $\mathcal{I}$, ($E$, $\mathcal{A}_E$)  [\textbf{Ours}] & \textbf{44.1} & \textbf{51}  \\
 wo/ $\mathcal{I}$  & 42.1   & 44      \\
 wo/ $\mathcal{I}$, ($E$, $\mathcal{A}_E$) [Object Caption]         & 39.9 &   43   \\\hline
\end{tabular}%
}
\caption{Multimodal EARL accuracy (\%) on the \mmee and \swig dataset. We compare three GVLM prompting strategies to understand the effect of various conditioning variables used in the multimodal prompt to GVLM for generating object descriptions. The experiment is performed in the one-shot setting. \looseness=-1}
\label{tab:prompting}
\end{table}

We report the results in Table \ref{tab:prompting} by prompting ChatGPT under the one-shot setting. We find that the performance of the \name{} framework with the object descriptions generated without any context of the event image and event details i.e., object caption performs the worst on both datasets (Row 3). We observe that providing the event details improves the performance on both datasets (Row 2 and Row 3). Finally, we observe that the best performance is achieved when the GVLM is provided with the event image and event information in its context where the accuracy gains are higher for the \swig dataset than the \mmee dataset. Our experiment reveals that we need to provide event-oriented information in the input prompt to generate event-centric object descriptions that will eventually contribute to improved multimodal EARL. Finally, we observe that our results are robust to the ordering perturbations in the predefined template to the LLM in Appendix \S \ref{sec:ordering}, which makes our approach flexible.

\section{Related Work}

\textbf{Event Extraction:} Most of the prior work has focused on extracting events and their structures grounded in the text modality such as text documents \cite{ahn2006stages,nguyen-etal-2016-joint-event,nguyen-grishman-2015-event,nguyen2019one,lin-etal-2020-joint,yang-mitchell-2016-joint,paolini2021structured,li2021document}. These works depend on training their models on the large-scale dataset with a complete event annotation \cite{doddington2004automatic,song2015light}. Previous studies have shown that these models do not generalize to unseen event types and new domains \cite{parekh2023geneva}. Other works in text-based event extraction have thus focused on training their models in a data-efficient manner \cite{lu2021text2event,hsu2022degree}. In this work, we utilize modern generative models that are capable of understanding the event task descriptions with image contexts and performing well on them, thus keeping our approach training-free.

\textbf{Multimodal Event Extraction:} While traditional event extraction has focused on text modality, multimodal event extraction is popularized by recent works \cite{yatskar2016situation,li2020cross,li2022clip}. These works focus on training weakly supervised representation models such as WASE and CLIP-Event on event annotated data \cite{doddington2004automatic,yatskar2016situation,li2020cross} to learn event semantics and structures. Similar to the previous works in text-based event extraction, these models will suffer from generalization issues with new event types. Additionally, there is no easy way to flexibly adapt them to new event types without further finetuning on expensive and hard-to-acquire event data. In this work, we show that \name{} generalizes to the unseen events by reasoning over the visual features of the image and object, and the event and its argument roles in its context.

\textbf{Generative Models:} Recently, there has been a surge of large-scale pretrained language models \cite{brown2020language,chatgpt,touvron2023llama,zhang2022opt,chowdhery2022palm,hoffmann2022training,lewis2019bart,raffel2020exploring} and multimodal generative models \cite{liu2023llava,li2022blip,li2023blip,dai2023instructblip,anas_awadalla_2023_7733589,Alayrac2022FlamingoAV,zhu2023minigpt,gong2023multimodalgpt}. These models have a remarkable ability to generalize to new tasks based on their descriptions in their context. \citet{wang2022language} use a combination of multimodal generative models and large language models to perform strong baselines on the few-shot video-language benchmarks. In our work, we leverage the flexibility and generalization capabilities of these models to perform well on multimodal EARL.
\section{Conclusion}

We propose \name{}, a training-free framework for multimodal event argument role labeling. 
Our experiments reveal that \name{} generalizes to various event types from the two datasets \mmee and \swig. This generalization capability is desirable for event extraction models since the existing methods rely on event-annotated data, which is expensive and time-consuming to acquire. Our work relies on the input prompts that follow the predefined human-crafted templates. Future work can focus on the automation of such templates based on the event information. Existing text-only and multimodal generative models \cite{bansal2022well,pratt2020grounded,nadeem2020stereoset} have been shown to encode harmful social biases. Future work should aim to contextualize social biases in multimodal EARL, and focus on reducing them.

\section{Acknowledgement}

Hritik Bansal and Po-Nien Kung are supported in part by AFOSR MURI grant FA9550-22-1-0380. In addition, this work was partially supported by Defense Advanced Research Project Agency (DARPA) grant HR00112290103/HR0011260656, CISCO and
ONR grant N00014-23-1-2780.

\section*{Limitations}

\name{} framework prompts modern vision-language generative models and large language models with event-specific task descriptions. These models have shown to encode harmful social biases and stereotypes that may be reflected in their output generations. For example, people with a specific color in an image might get labeled as `attackers' or `offenders' without understanding their true role in the depicted event. Future work should focus on the study of these generated object descriptions from GVLM and EARL predictions from the LLMs through human evaluation.

In our experiments, we perform the human assessment by employing mechanical turkers. All the labelers belonged to the US and thus will limit our assessment due to their own perceptual biases when presented with an event depicting an image and the participant objects. To obtain more reliable annotations, we can involve annotators from diverse regions.

Multimodal event extraction is a relatively new endeavor in comparison to traditional text-based event extraction methods. Thus, there are very few evaluation datasets to assess the multimodal models for their generalization capabilities to diverse domains and event types. It will be imperative to benchmark our approach on such diverse datasets for more comprehensive evaluation.



\bibliography{anthology,custom}
\bibliographystyle{acl_natbib}
\newpage
\appendix


\section{Multimodal EARL}
\label{app:method_earl}

Firstly, we assume that a given image has a single major event associated with it since there can be no event or multiple events depicted in the image. Secondly, we assume that every object (bounding box) participates in the event and hence has an associated event argument role label. However, we do incorporate the ability to predict `Other' if the model does find a good match in the candidate argument role labels. Thirdly, we allow multiple objects in a given image to have the same event argument role label, as shown in Figure \ref{fig:example}. 

Existing methods \cite{li2020cross,li2022clip} train the models on high-quality unimodal or multimodal data annotated with the events, entities, and their relations. For example, ACE is a text-based dataset, imSitu \cite{yatskar2016situation} is for actions (or events) in the images, and \cite{li2022clip} utilizes VOA news as a source for multimedia events. In practice, collecting and annotating these datasets is time-consuming and expensive. Furthermore, the inability of the existing models to adapt to new event types and new domains without specialized training data makes them less generalizable.

\section{Setup-Baseline}
\label{app:baseline}




\subsection{Supervised Setting}
\label{app:clip_supervised}


We train our model with a batch size = 32, epochs = 5, AdamW optimizer \cite{loshchilov2017decoupled} with a linear warmup of 100 steps followed by cosine annealing. We perform a hyperparameter search over three learning rates = $\{1e-4, 1e-5, 1e-6\}$. We find that the model trained with $1e-5$ performs the best on the test examples of the \swig dataset during inference. Each finetuning experiment took $4-5$ hours on a single A6000 48GB Nvidia GPU.

Given the lack of available training data for the \mmee domain, we conduct an evaluation of CLIP (+\swig) as a transfer learning setup. Our findings reveal that the finetuned CLIP (+\swig) model, utilizing a learning rate of $1e-4$, demonstrates the most optimal performance on the \mmee dataset.

\section{Precision, Recall, and F1 Metrics}
\label{sec:p_r_f1}

In Table \ref{app_table:prf1}, we report the precision, recall and F1 of GenEARL, Zero-shot CLIP baseline, and Supervised CLIP finetuned on the \mmee and \swig dataset. We find that the trends (rankings) of the multimodal EARL methodologies are the same as reported in Table \ref{tab:main_earl_results}. Specifically, we observe that GenEARL (3-shot) outperforms Supervised CLIP under the transfer learning settings on \mmee. In addition, we find that the GenEARL (0-shot and 3-shot) outperform the zero-shot CLIP baseline on the \swig dataset.

\begin{table*}[h]
\begin{tabular}{lccc|ccc}
\hline
\multirow{2}{*}{\textbf{Method}}       & \multicolumn{3}{c|}{\textbf{\mmee}}    & \multicolumn{3}{c}{\textbf{\swig}}    \\
                              & \textbf{Precision} & \textbf{Recall} & \textbf{F1}     & \textbf{Precision} & \textbf{Recall} & \textbf{F1}     \\\hline
Zero-shot CLIP-B/32                 & 41.6\%    & 28\%   & 33.2\% & 22\%      & 27.3\% & 24.4\% \\
GenEARL (0-shot)              & 65.3\%    & 38.4\% & 48.4\% & 36.5\%    & 47\%   & 41\%   \\
GenEARL (3-shot)              & 66.1\%    & 44.7\% & 53.3\% & 47\%      & 55\%   & 50.7\% \\
\textcolor{gray}{Supervised CLIP-B/32 (+SwiG)} & \textcolor{gray}{66\%}      & \textcolor{gray}{40.3\%} & \textcolor{gray}{50\%}   & \textcolor{gray}{61.5\%}    & \textcolor{gray}{72\%}   & \textcolor{gray}{66\%}  \\\hline
\end{tabular}
\caption{Precision, Recall, and F1 scores for Multimodal Event Argument Role Labeling on \mmee and \swig datasets.}
\label{app_table:prf1}
\end{table*}

\section{Effect of ordering perturbation in the predefined template}
\label{sec:ordering}

In \S \ref{method:llm}, we describe a predefined template $\mathcal{T}_\ell$ that converts the input variables (object role description, event type, image description, possible argument roles and their definitions) from the generative vision language model to detailed task description for LLM. In our experiments, we find that the LLM (ChatGPT) predictions are not sensitive to the order in which the input variables are presented to it in the predefined template. Specifically, we observe a slight change of $\pm 2\%$ for \name{} (3-shot) on both datasets over different orderings of the input variables to the LLM template. This indicates that the predefined template is very flexible and can easily generalize to new datasets.

\section{Using Alpaca for EARL Prediction}
\label{app:alpaca_earl}

We evaluate the probability scores from the Alpaca model $\mathcal{M}$ over the possible event argument roles $a_j \in \mathcal{A}_E$ given by $p_{\mathcal{M}}(a_j | \mathcal{C})$ where $p_{\mathcal{M}}(. | \mathcal{C})$ is the learned probability distribution of the Alpaca model, and $\mathcal{C}$ is the context described in \S \ref{method:llm}. The predicted EARL $\tilde{a}_i$ for the object $o_i$ is the one that maximizes the probability $\tilde{a}_i = \text{argmax}_{j}p_{\mathcal{M}}(a_j | \mathcal{C})$. We provide the EARL prediction template used for Alpaca experiments in the zero-shot settings in Table \ref{tab:alpaca_main}.

\section{Using LLaVA for EARL Prediction}
\label{app:llava_earl}

\begin{figure}
\centering
\resizebox{\linewidth}{!}{
\begin{tabular}{p{\linewidth}}
\toprule
\textbf{Input}: Image $\mathcal{I}$, Object $o$, Event $E$, Event argument role labels (and definitions) $\mathcal{A}$\\
\midrule
\textbf{Prompt:} \\
Image 1: 256 $\times <\mathcal{I}$ visual feature tokens$>$ \\
Image 2: 256 $\times <o$ visual feature tokens$>$ \\
What is the role of the entity in the "Image 2" in the context of the $E$ event in the "Image 1"? The possible argument roles of the objects in the $E$ event include $\mathcal{A}$. Choose only one of these options.\\
\bottomrule
\end{tabular}}
\caption{Multimodal input template used to prompt the GLVM for event argument role labeling. Here, the model is provided with the image, event and possible argument role labels (and definitions). We add `Other' to the list of possible argument role labels in case it does not prefer any of the existing event argument role labels. We get 256 visual tokens for the image or the object by projecting the raw input into the vision embedding space using the visual processing module of the GVLM.}
\label{tab:llava_only}
\end{figure}

To this end, we evaluate the probability scores from the LLaVA model $\mathcal{G}$ over the possible event argument roles $a_j \in \mathcal{A}_E$ given by $p_{\mathcal{G}}(a_j | \mathcal{P})$ where $p_{\mathcal{M}}(. | \mathcal{P})$ is the learned probability distribution of the LLaVA model, and $\mathcal{P}$ is the context described in \S \ref{method:gvlm}. The predicted EARL $\tilde{a}_i$ for the object $o_i$ is the one that maximizes the probability $\tilde{a}_i = \text{argmax}_{j}p_{\mathcal{G}}(a_j | \mathcal{P})$. We provide the EARL prediction template used for LLaVA in the zero-shot settings in Figure \ref{tab:llava_only}.

\section{Input Prompts for LLaVA}

We provide the multimodal input templates used for prompting LLaVA in the \name{} framework. By default, we use the template presented in the Figure \ref{tab:llava_main}. We use the templates Figure \ref{tab:llava_wo_image} and Figure \ref{tab:llava_wo_image_event} for the experiments in \S \ref{exp:earl_gvlm}. Since LlaVA is known to have a limited context token length of 2048 and like most transformer architectures \cite{tay2020long} is susceptible to suffer from ignoring long-range dependencies, we create object role descriptions for one object $o_i$ at a time. 

\section{Event Types}

To maintain the argument role density, we filter the event types with less than three argument roles from the \mmee dataset and the \swig dataset. We provide the list of event types included in our multimodal EARL evaluation in Table \ref{m2e2_event_list} and Table \ref{swig_event_list}.

\begin{table*}[h]
\begin{center}
\resizebox{\linewidth}{!}{
\begin{tabular}{ll}
\hline
     \textbf{Event Types}&             \textbf{Event Argument Role Definition}                                                                                                     \\\hline
Life: Die                  & {[}Victim{]} died at place, or was killed by {[}Agent{]} using {[}Instrument{]}                                  \\
Movement.Transport         & {[}Agent{]} transported {[}Artifact or Person{]} in {[}Instrument{]} from {[}Origin{]} to {[}Destination{]}      \\
Conflict.Attack            & {[}Attacker{]} attacked or assaulted {[}Target{]} {[}Instrument{]} at place                                      \\
Conflict.Demonstrate       & {[}Demonstrator{]} was in a demonstration at place holding {[}Instrument{]} and supervised by {[}Police{]} \\
Justice.Arrest Jail        & {[}Agent{]} arrested or jailed {[}Person{]} using {[}Instrument{]} at place                                      \\
Transaction.Transfer Money & {[}Giver{]} gave {[}Money{]} to {[}Recipient{]}                                                             \\\hline
\end{tabular}
}
\caption{The list of six event types from the \mmee dataset included in our experiments. The argument role definitions are taken from \cite{li2020cross}.}
\label{m2e2_event_list}
\end{center}
\end{table*}

\begin{table*}[h]
\begin{center}
\resizebox{\linewidth}{!}{
\begin{tabular}{p{\linewidth}}
\hline
`tattooing', 'splashing', 'emerging', 'pasting', 'inflating', 'displaying', 'feeding', 'shredding', 'flicking', 'shaving', 'flinging', 'yanking', 'putting', 'decorating', 'mashing', 'baking', 'moistening', 'inserting', 'chopping', 'parachuting', 'giving', 'cooking', 'falling', 'ramming', 'pushing', 'wrapping', 'leaping', 'packing', 'smearing', 'whisking', 'sharpening', 'sealing', 'examining', 'raking', 'punching', 'whipping', 'watering', 'steering', 'distributing', 'spying', 'uncorking', 'squeezing', 'pawing', 'dripping', 'carrying', 'pinching', 'tying', 'arranging', 'tearing', 'blocking', 'practicing', 'potting', 'installing', 'dousing', 'making', 'unpacking', 'mowing', 'stacking', 'stinging', 'heaving', 'pulling', 'breaking', 'docking', 'destroying', 'drying', 'poking', 'tickling', 'tilting', 'hoisting', 'nailing', 'lighting', 'shelving', 'wagging', 'lifting', 'surfing', 'gluing', 'hauling', 'erasing', 'bathing', 'serving', 'drinking', 'soaking', 'writing', 'coaching', 'flipping', 'slicing', 'educating', 'bothering', 'sweeping', 'washing', 'spinning', 'dyeing', 'filling', 'shaking', 'signing', 'catching', 'scratching', 'shearing', 'ejecting', 'peeling', 'hanging', 'sliding', 'igniting', 'pumping', 'clenching', 'fastening', 'repairing', 'jumping', 'striking', 'manicuring', 'farming', 'photographing', 'mining', 'selling', 'bouncing', 'leaking', 'autographing', 'scraping', 'providing', 'tipping', 'patting', 'interrogating', 'emptying', 'hurling', 'scooping', 'sewing', 'massaging', 'attacking', 'signaling', 'dissecting', 'spanking', 'hugging', 'crashing', 'coloring', 'scrubbing', 'rinsing', 'fueling', 'begging', 'spraying', 'frying', 'stitching', 'spilling', 'covering', 'dipping', 'wheeling', 'constructing', 'burying', 'throwing', 'drenching', 'injecting', 'pricking', 'clearing', 'plummeting', 'wetting', 'weighing', 'deflecting', 'opening', 'lapping', 'vacuuming', 'kicking', 'picking', 'folding', 'operating', 'smashing', 'stroking', 'pouring', 'tuning', 'dusting', 'attaching', 'dropping', 'resting', 'rocking', 'paying', 'guarding', 'leaning', 'kissing', 'slapping', 'harvesting', 'piloting', 'welding', 'stuffing', 'rubbing', 'buttering', 'pruning', 'pinning', 'dragging', 'fetching', 'locking', 'shoveling', 'cramming', 'disciplining', 'planting', 'cleaning', 'spitting', 'climbing', 'caressing', 'flexing', 'drawing', 'assembling', 'lathering', 'mending', 'loading', 'combing', 'unloading', 'trimming', 'molding', 'checking', 'descending', 'aiming', 'performing', 'fishing', 'taping', 'immersing', 'ailing', 'sketching', 'grinding', 'filming', 'measuring', 'eating', 'buckling', 'submerging', 'helping', 'crafting', 'drumming', 'plunging', 'carting', 'curling', 'spreading', 'clipping', 'carving', 'hitting', 'retrieving', 'placing', 'launching', 'crushing', 'pitching', 'offering', 'strapping', 'adjusting', 'moisturizing', 'tasting', 'wiping', 'sprinkling', 'buying', 'applying', 'stapling', 'stirring', 'floating', 'shooting', 'microwaving', 'fixing', 'milking', 'brushing', 'vaulting', 'unlocking', 'extinguishing', 'building', 'painting', 'waxing', 'stripping', 'prying', 'crowning'
\\\hline

\end{tabular}
}
\caption{The list of 262 event types from the \swig dataset included in our experiments.}
\label{swig_event_list}
\end{center}
\end{table*}

\section{Qualitative Example}
\label{app:qualitative}

\begin{figure*}[h]
    \centering
    \includegraphics[width=\linewidth]{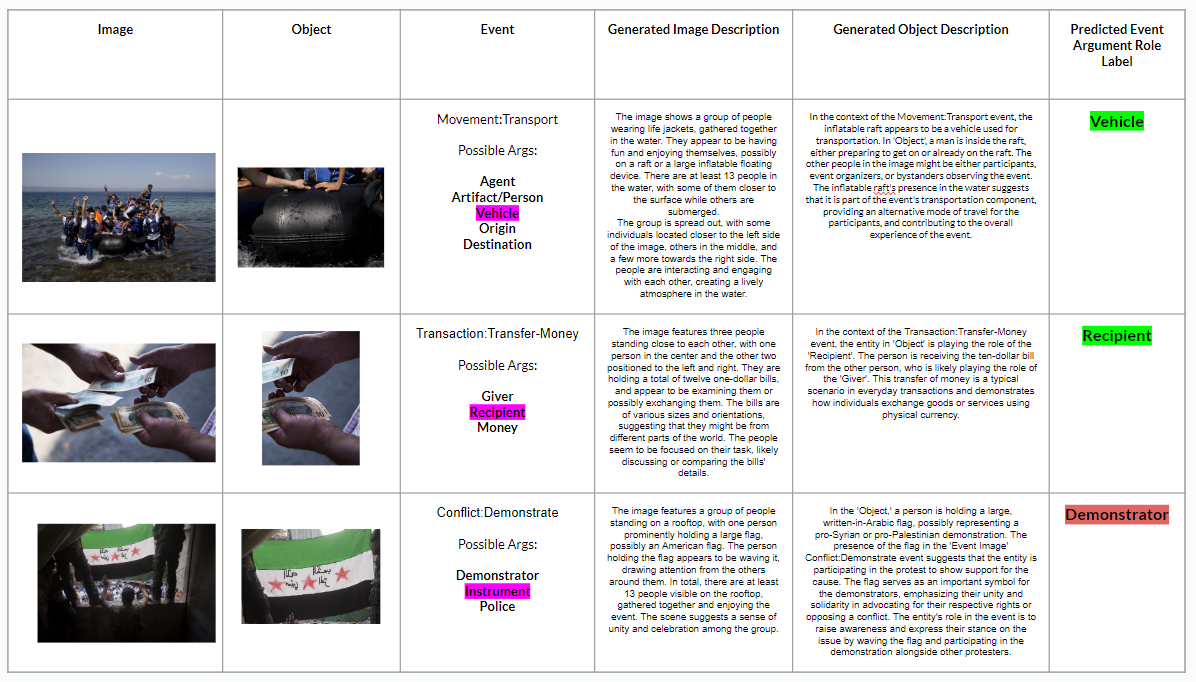}
    \caption{Qualitative Example for the predictions by the \name{} framework in three-shot setting.}
    \label{fig:qual_example}
\end{figure*}

We provide three qualitative examples from the \mmee dataset in Figure \ref{fig:qual_example}.

\section{Human Assessment Screenshot}
\label{app:turker}

\begin{figure*}[h]
    \centering
    \includegraphics[width=\linewidth]{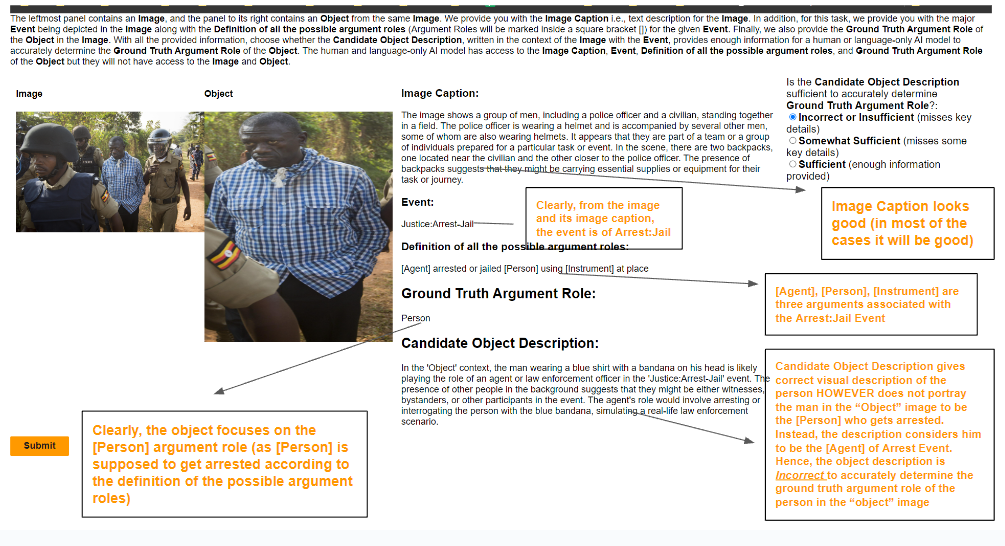}
    \caption{Screenshot of a solved example shown to the human labelers before they attempt the qualification test.}
    \label{fig:turker_solved}
\end{figure*}

\begin{figure*}[h]
    \centering
    \includegraphics[width=\linewidth]{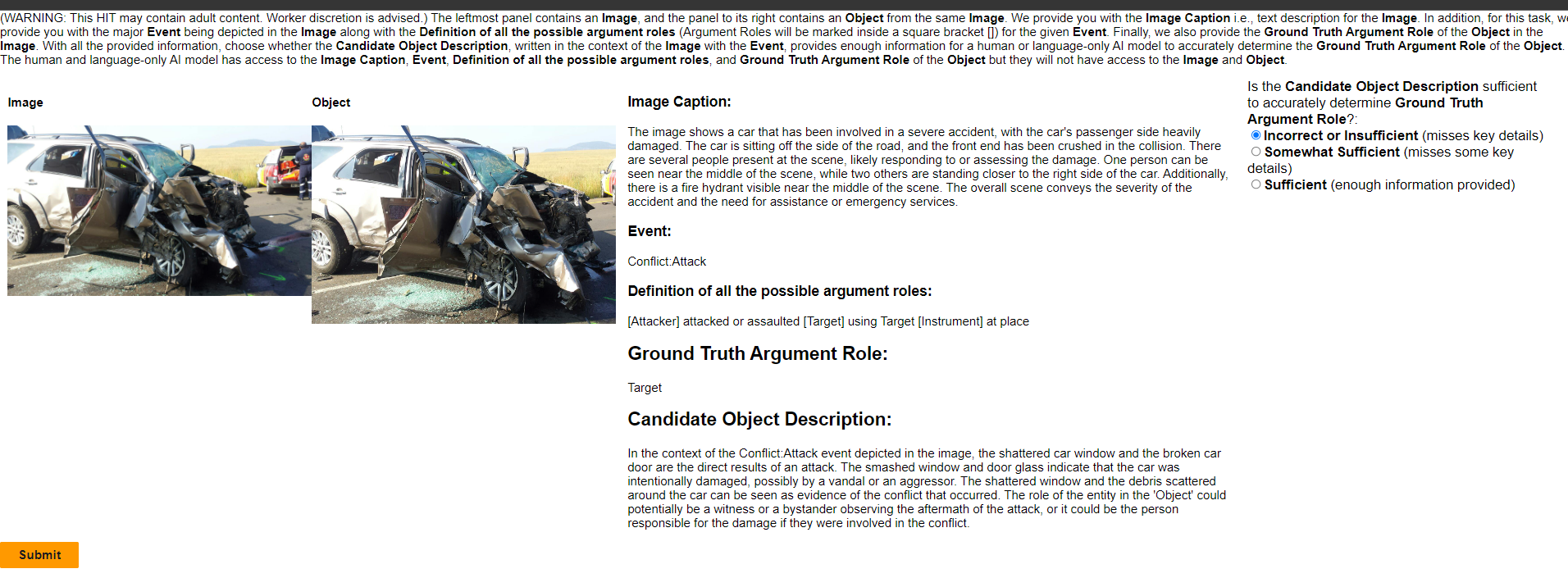}
    \caption{Screenshot of an unsolved example shown to the human labelers during the actual annotation.}
    \label{fig:turker_unsolved}
\end{figure*}

Here, we provide the screenshots for our human assessments. Figure \ref{fig:turker_solved} illustrates a solved example shown to the labelers before the qualification test to make them understand the task. Figure \ref{fig:turker_unsolved} illustrates an unsolved example assigned to 3 labelers during the actual annotation.

\section{Can two-stage generative prompting perform multimodal event detection?}

In this work, we present \name{}, a combination of prompting a generative vision-language model followed by a large language model for multimodal EARL. Here, we aim to study whether the same principles can be adapted to perform multimodal event detection i.e., classifying the given image into one of the event types from the dataset. 

To this end, we first generate an image caption from LLaVA-7B. Then, we prompt ChatGPT to predict the event type based on the generated image description and the list of possible event types (and their definitions). We perform this experiment in a zero-shot setting. Further, we compare our model against the zero-shot pretrained CLIP \cite{radford2019language}, and zero-shot and supervised CLIP-Event \cite{li2022clip}. We report the results in Table \ref{app_exp:ed}.

\begin{figure}
    \centering
    \includegraphics[scale = 1]{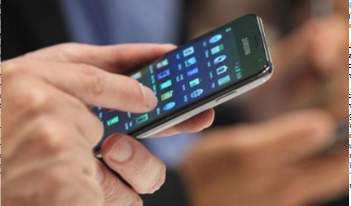}
    \caption{An example image depicting `dialing' action from the \swig dataset.}
    \label{fig:swig_llava}
\end{figure}

We find that the two-stage generative prompting framework outperforms all the other methods including the supervised baselines for multimodal event detection on the \mmee dataset. This suggests that generative prompting is a flexible and generalizable technique that can be used for event detection without access to any event-annotated training data. Surprisingly, we find that the two-stage framework performs poorly on the \swig dataset. We perform a manual inspection of several examples where the predicted action does not match the ground action. Our investigation reveals that the two-stage generative prompting does provide reasonable predictions for the actions depicted in the images. For example, as shown in Figure \ref{fig:swig_llava}, the ground-truth action annotated in the \swig dataset is `dialing'. However, the two-stage setup predicts `pressing' and `navigating' actions which are legitimate through visual inspection of Figure \ref{fig:swig_llava}. Although we do not conduct a large-scale human evaluation of our approach, we attribute the low performance of the two-stage generative feedback on \swig dataset to its restricted ground-truth action verb assignments (`dialing') despite the presence of other related actions in the list of event types (`pressing', `typing'). Although not shown here, we find that providing few-shot examples to ChatGPT does not improve the performance on the \swig dataset.

\begin{table}[h]
\begin{center}
\resizebox{\linewidth}{!}{
\begin{tabular}{lcc}
\hline
Method                & \mmee Acc. (\%)  & \swig Acc. (\%) \\\hline
Zero-shot-LLaVA-ChatGPT  & \textbf{74.3} & 11   \\
Zero-shot-CLIP (B/32)& 65.7  & 28.3 \\
Zero-shot-CLIP-Event & 70.8  & \textbf{31.4} \\\hline
\textcolor{gray}{Supervised-CLIP-Event} & \textcolor{gray}{72.8}  & \textcolor{gray}{45.6}\\\hline
\end{tabular}
}
\caption{Event Detection Accuracy (\%) of generative prompting (LLaVA followed by ChatGPT), pretrained CLIP, and CLIP-Event \cite{li2022clip} models on the \mmee and \swig dataset.}
\label{app_exp:ed}
\end{center}
\end{table}

\begin{figure*}
    \centering
    \includegraphics[width = \linewidth]{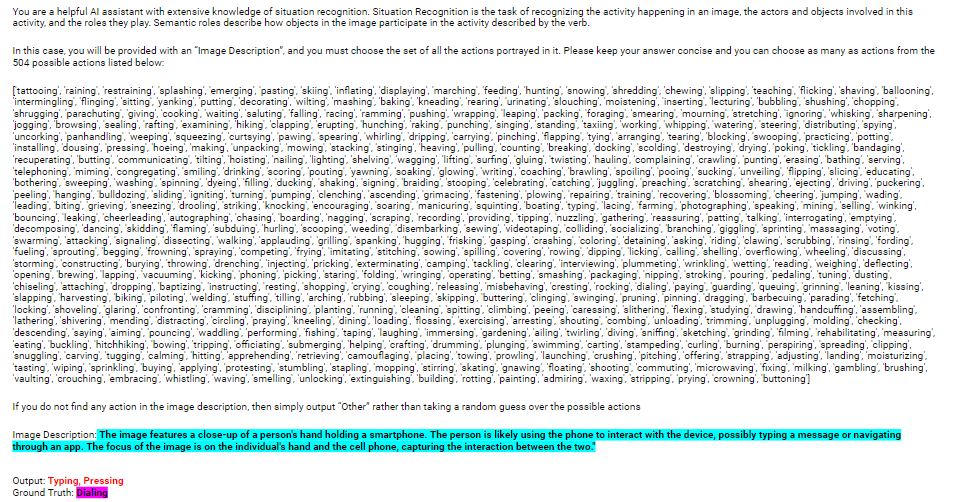}
    \caption{Example prompt for predicting the event type using a two-stage generative framework using the \swig dataset. We highlight the generated caption for image (Figure \ref{fig:swig_llava}) in blue. The ChatGPT outputs are bolded in red while the ground truth action is bolded in pink.}
    \label{fig:swig_ed}
\end{figure*}

\begin{figure}
\centering
\resizebox{\linewidth}{!}{
\begin{tabular}{p{\linewidth}}
\toprule
\textbf{Input}: Image $\mathcal{I}$, Object $o$, Event $E$, Event argument role labels (and definitions) $\mathcal{A}$\\
\midrule
\textbf{Prompt:} \\
Image 1: 256 $\times <\mathcal{I}$ visual feature tokens$>$ \\
Image 2: 256 $\times <o$ visual feature tokens$>$ \\
Describe the role of the entity in ``Image 2'' in the context of the $E$ event in ``Image 1''. The possible argument roles of the objects performing $E$ event include $\mathcal{A}$. Please be concise with your answer. \\
\bottomrule
\end{tabular}}
\caption{Multimodal input template used to prompt the GLVM in the \name{} framework. Here, the model is provided with the image, event and possible argument role labels (and definitions). We get 256 visual tokens for the image or the object by projecting the raw input into the vision embedding space using the visual processing module of the GVLM.}
\label{tab:llava_main}
\end{figure}

\begin{figure}
\centering
\resizebox{\linewidth}{!}{
\begin{tabular}{p{\linewidth}}
\toprule
\textbf{Input}: Object $o$, Event $E$, Event argument roles (and definitions) $\mathcal{A}$\\
\midrule
\textbf{Prompt:} \\
Image: 256 $\times <o$ visual feature tokens$>$ \\ \\
Describe the role of the entity in ``Image'' in the context of the $E$ event. The possible argument roles of the objects performing $E$ event include $\mathcal{A}$. Please be concise with your answer. \\
\bottomrule
\end{tabular}}
\caption{Multimodal input template used to prompt the GLVM in the \name{} framework. Here, the model is provided with the event and possible argument role labels (and definitions). We get 256 visual tokens for the object by projecting the raw input into the vision embedding space using the visual processing module of the GVLM.}
\label{tab:llava_wo_image}
\end{figure}

\begin{figure}
\centering
\resizebox{\linewidth}{!}{
\begin{tabular}{p{\linewidth}}
\toprule
\textbf{Input}: Object $o$\\
\midrule
\textbf{Prompt:} \\
Image: 256 $\times <o$ visual feature tokens$>$ \\\\
Describe the ``Image'' concisely.\\
\bottomrule
\end{tabular}}
\caption{Multimodal input template used to prompt the GLVM in the \name{} framework. Here, the model is provided just the event participant object. We get 256 visual tokens for the object by projecting the raw input into the vision embedding space using the visual processing module of the GVLM.}
\label{tab:llava_wo_image_event}
\end{figure}


\begin{figure*}
\centering
\resizebox{\linewidth}{!}{
\begin{tabular}{p{\linewidth}}
\toprule
\textbf{Input}: Image Caption $\mathcal{I}$, Generated object descriptions $\{g_1, g_2,\ldots,g_k\}$ from LLaVA for the $k$ participants objects in the image , Event $E$, Event argument role labels (and definitions) $\mathcal{A}$\\
\textbf{Setting}: Zero-shot\\
\midrule
\textbf{Prompt:}\\
You are a helpful AI assistant with extensive knowledge of event argument extraction. Worldwide events are documented in raw text on various online platforms, and it is crucial to extract useful and concise information about them for downstream applications. \\\\

In this case, you will be provided with an “Event Image Description”, the “Event” portrayed in the image, a generic “Event Argument Roles Definition” that helps you to understand the argument roles that are grounded in different objects in the image, and the “Object Role” descriptions to describe the role of specific objects in the context of the "Event Image". Based on the provided information, you need to tell the argument roles associated with different objects in the image. \\\\

Remember that:
1) The number of possible event argument roles can sometimes be equal to, more, or less than the number of objects detected for the "Event Image". \\
2) Multiple objects may get identical event argument roles, but not always.\\
3) It is completely possible that some of the event argument roles are not grounded in any of the objects detected for the "Event Image".\\

Please keep your answer concise. You can choose to assign a "Other" argument role if you are not sure about the argument role for a particular object.\\\\

Event Image Description: $\mathcal{I}$\\\\

Event: $E$ \\\\

Event Argument Role Definition: $\mathcal{A}$ \\\\

Role of Object 1: $g_1$\\
.\\
.\\
Role of Object k: $g_k$\\\\

Argument Role of Object 1: \\
.\\
.\\
Argument Role of Object k: \\
\midrule
\textbf{Output:}\\
Argument Role of Object 1: $\hat{a}_1$\\
.\\
.\\
Argument Role of Object k: $\hat{a}_k$\\

\bottomrule
\end{tabular}}
\caption{LLM template used to prompt the ChatGPT in the \name{} framework. This template is utilized under the zero-shot settings. Due to cost considerations associated with prompting ChatGPT, we perform multimodal EARL in a batch of all the event participant objects in an image.}
\label{tab:chatgpt_main}
\end{figure*}

\begin{figure*}
\centering
\resizebox{.95\linewidth}{!}{
\begin{tabular}{p{\linewidth}}
\toprule
\textbf{Solved Input}: Image Caption $\mathcal{I}_s$, Generated object descriptions $\{g_{s,1}, g_{s,2},\ldots,g_{s,l}\}$ from LLaVA for the $l$ participants objects in the image , Event $E_s$, Event argument role labels (and definitions) $\mathcal{A}_s$\\
\textbf{Query Input}: Image Caption $\mathcal{I}$, Generated object descriptions $\{g_1, g_2,\ldots,g_k\}$ from LLaVA for the $k$ participants objects in the image , Event $E$, Event argument role labels (and definitions) $\mathcal{A}$\\
\textbf{Setting}: One-shot\\
\midrule
\textbf{Prompt:}\\
You are a helpful AI assistant with extensive knowledge of event argument extraction. Worldwide events are documented in raw text on various online platforms, and it is crucial to extract useful and concise information about them for downstream applications. \\

In this case, you will be provided with an “Event Image Description”, the “Event” portrayed in the image, a generic “Event Argument Roles Definition” that helps you to understand the argument roles that are grounded in different objects in the image, and the “Object Role” descriptions to describe the role of specific objects in the context of the "Event Image". Based on the provided information, you need to tell the argument roles associated with different objects in the image. \\

Remember that:
1) The number of possible event argument roles can sometimes be equal to, more, or less than the number of objects detected for the "Event Image". \\
2) Multiple objects may get identical event argument roles, but not always.\\
3) It is completely possible that some of the event argument roles are not grounded in any of the objects detected for the "Event Image".\\

Please keep your answer concise. You can choose to assign a "Other" argument role if you are not sure about the argument role for a particular object.\\

We will first show a single solved instance of the task, and then you will complete the task on a new query.\\\\

Solved Instance:\\

Event Image Description: $\mathcal{I}_s$\\

Event: $E_s$\\

Event Argument Role Description: $\mathcal{A}_s$\\

Role of Object 1: $g_{s,1}$\\
.\\
.\\
Role of Object l: $g_{s,l}$\\

Argument Role of Object 1: $a_{s,1}$\\
.\\
.\\
Argument Role of Object l: $a_{s,l}$\\\\

Query Instance: \\\\

Event Image Description: $\mathcal{I}$\\

Event: $E$ \\

Event Argument Role Definition: $\mathcal{A}$ \\

Role of Object 1: $g_1$\\
.\\
.\\
Role of Object k: $g_k$\\

Argument Role of Object 1: \\
.\\
.\\
Argument Role of Object k: \\
\midrule
\textbf{Output:}\\
Argument Role of Object 1: $\hat{a}_1$\\
.\\
.\\
Argument Role of Object k: $\hat{a}_k$\\

\bottomrule
\end{tabular}}
\caption{LLM template used to prompt the ChatGPT in the \name{} framework. This template is utilized under the one-shot settings.}
\label{tab:chatgpt_one_shot}
\end{figure*}

\begin{figure*}
\centering
\resizebox{\linewidth}{!}{
\begin{tabular}{p{\linewidth}}
\toprule
\textbf{Input}: Image Caption $\mathcal{I}$, Generated object description $g$ from LLaVA for the one of the participant objects in the image , Event $E$, Event argument role labels (and definitions) $\mathcal{A}$\\
\textbf{Setting}: Zero-shot\\
\midrule
\textbf{Prompt:}\\

Given an “Event Image Description”, the “Event” portrayed in the image, a generic “Event Argument Roles Definition” that helps you to understand the argument roles that are grounded in different objects in the image, and the “Object Role” description to describe the role of the object in the context of the "Event Image". Based on the provided information, you need to tell the argument roles associated with different objects in the image. You can choose to assign an "Other" argument role if you are not sure about the argument role for a particular object. \\\\

Event Image Description: $\mathcal{I}$\\\\

Event: $E$ \\\\

Event Argument Role Definition: $\mathcal{A}$ \\\\

Role of Object: $g$\\\\

Argument Role of Object: \\
\bottomrule
\end{tabular}}
\caption{LLM template used to prompt the Alpaca-7B. This template is utilized under the zero-shot settings. Due to the context length limitations, we label one participating object at a time.}
\label{tab:alpaca_main}
\end{figure*}

\end{document}